\documentclass[conference]{ieeeconf}  % Comment this line out if you need a4paper

\IEEEoverridecommandlockouts                              % This command is only needed if 
\usepackage{amsmath} % assumes amsmath package installed
\usepackage{amssymb}  % assumes amsmath package installed
\usepackage{cite}
\usepackage{graphicx}
\usepackage{subcaption}
\usepackage[flushleft]{threeparttable}
\usepackage{booktabs}
\usepackage{multirow}
\usepackage{color, colortbl}
\usepackage{balance}
\usepackage{xcolor}
\usepackage{url} 
\usepackage{hyperref}
\usepackage{tabularx}
\usepackage{makecell}
\usepackage{graphicx}
\usepackage{subcaption}

\title{\LARGE \bf
LSRE: Latent Semantic Rule Encoding for Real-Time Semantic Risk Detection in Autonomous Driving 
}
\author{Qian Cheng, Weitao Zhou, Cheng Jing, Nanshan Deng, Junze Wen, Zhaoyang Liu, Kun Jiang, Diange Yang 
\thanks{Qian Cheng, Weitao Zhou, Cheng Jing, Nanshan Deng, Junze Wen, Zhaoyang Liu, Kun Jiang,  and Diange Yang are with the School of Vehicle and Mobility, Tsinghua University.} 
\thanks{Corresponding to Weitao Zhou, Diange Yang  (zhouwt@tsinghua.edu.cn, ydg@tsinghua.edu.cn)}
}

\begin{document}

\maketitle
\thispagestyle{empty}
\pagestyle{empty}

\begin{abstract}

Real-world autonomous driving must adhere to complex human social rules that extend beyond legally codified traffic regulations. Many of these semantic constraints—such as yielding to emergency vehicles, complying with traffic officers’ gestures, or stopping for school buses—are intuitive for humans yet difficult to encode explicitly. Although large vision–language models (VLMs) can interpret such semantics, their inference cost makes them impractical for real-time deployment.
This work proposes LSRE, a Latent Semantic Rule Encoding framework that converts sparsely sampled VLM judgments into decision boundaries within the latent space of a recurrent world model. By encoding language-defined safety semantics into a lightweight latent classifier, LSRE enables real-time semantic risk assessment at 10 Hz without per-frame VLM queries.
Experiments on six semantic-failure scenarios in CARLA demonstrate that LSRE attains semantic risk detection accuracy comparable to a large VLM baseline, while providing substantially earlier hazard anticipation and maintaining low computational latency. LSRE further generalizes to rarely seen semantic-similar test cases, indicating that language-guided latent classification offers an effective and deployable mechanism for semantic safety monitoring in autonomous driving.

\end{abstract}

\section{Introduction}

Autonomous driving in open-world environments requires more than accurate perception and robust control—it demands adherence to nuanced human social rules~\cite{shalev2017formal}. Many such semantics, including yielding to emergency vehicles, obeying traffic officers over traffic lights, or interpreting temporary construction-zone layouts, are intuitive for humans yet difficult to encode as explicit rules~\cite{kuefler2017imitating}. These context-dependent constraints form a class of \textit{semantic safety requirements} that conventional rule-based or geometric methods cannot capture, but whose violations frequently lead to critical long-tail failures\cite{zhou2022dynamically}. For example, human drivers coordinate through social cues and implicit norms beyond formal traffic regulations—autonomous vehicles strictly following only traffic
rules may struggle with human-like negotiation in complex traffic~\cite{amodei2016concrete}.  Similarly, while traffic-law encodings such as reachability or temporal logic cover structured interactions, they fall short in representing latent social semantics that emerge only through interactive context ~\cite{Sadigh-RSS-16}.

% Existing safety mechanisms primarily rely on traffic-rule logic, heuristic filtering, or reachability-based reasoning. While effective for geometric safety (e.g., collision avoidance, lane boundaries), they fundamentally lack the ability to handle human-defined semantics. Recent vision–language models (VLMs) provide strong high-level understanding and offer a promising avenue for semantic safety validation. However, direct per-frame VLM reasoning is computationally prohibitive for real-time driving, and its outputs lack temporal consistency. As a result, current systems either ignore these semantic constraints or apply VLMs offline, leaving a gap between semantic understanding and deployable safety. 、

Existing safety mechanisms primarily rely on traffic-rule logic, heuristic filtering, or reachability-based reasoning\cite{althoff2018commonroad}. While these methods are effective for geometric safety---such as maintaining lane boundaries, enforcing collision-free trajectories, and ensuring kinematic feasibility---they fundamentally lack the ability to capture human-defined semantics that emerge from social context or temporary traffic configurations. Prior work on formal safety verification and reachability analysis \cite{ivanovic2023verification} demonstrates strong guarantees for physical constraints, yet these approaches cannot express high-level obligations such as yielding to emergency vehicles or obeying traffic officers.

Recent vision--language models (VLMs) offer a promising direction by providing rich high-level understanding of road semantics \cite{kim2023lavit}. VLMs can identify nuanced traffic cues that are otherwise absent from explicit rule sets or HD maps\cite{kar2023llm4drive}. However, direct per-frame VLM reasoning remains computationally prohibitive for real-time driving---typical inference times exceed 200--800~ms per frame \cite{zhou2024vision}. Also, VLM outputs lack temporal consistency due to the absence of predictive structure. Consequently, current systems either ignore these semantic constraints or apply VLMs only in offline analysis pipelines, leaving a persistent gap between semantic understanding and deployable safety mechanisms in autonomous driving.

% \begin{figure}
%     \centering
%     \includegraphics[width=1\linewidth]{images/idea.jpg}
%    \caption{  VLM-guided latent risk detection}
%     \label{fig:idea}
% \end{figure}

To bridge this gap, we propose \textbf{LSRE (Latent Semantic Rule Encoding)}, a framework that distills language-defined semantic safety rules into a lightweight classifier operating inside the latent space of a recurrent world model. LSRE queries a VLM sparsely during training to obtain semantic risk labels, then encodes these labels as decision boundaries in the latent dynamics space. During deployment, the latent classifier provides frame-level semantic risk predictions at 10 Hz without per-frame VLM inference, enabling real-time enforcement of human-understandable social semantics. 

\begin{itemize}

\item Latent Semantic Rule Encoding:
We propose \textit{LSRE}, which distills language-defined semantic rules into decision boundaries in the latent space of a recurrent world model, enabling real-time semantic safety assessment without per-frame VLM inference.

\item VLM-Supervised Latent Classifier with Temporal Anticipation:
We design a lightweight latent classifier trained under sparse VLM supervision and enhanced with short-horizon latent rollouts and hysteresis-based filtering. This combination provides both stable predictions and early hazard anticipation at millisecond-level latency.

\item Semantic-Failure Benchmark and Evaluation:
We construct a CARLA benchmark with six semantic-failure scenario variants. LSRE matches VLM-level accuracy, detects hazards substantially earlier, and generalizes to semantic-similar but unseen scenes.
\end{itemize}

% \vspace{2mm}
% \noindent
% \textbf{Overview.}
% An overview of the proposed framework is shown in figure ~\ref{fig:idea}, which illustrates three representative paradigms for risk identification.

% (a) The first paradigm is perception and prediction-based risk detection, where predefined rules are applied to recognize potentially unsafe behaviors. Although this approach offers basic spatial–temporal reasoning and requires relatively low computational cost, it lacks the ability to capture semantic failure cases.

% (b) The second paradigm performs risk detection directly using large vision-language models. These models provide strong semantic understanding, but their inference is computationally expensive and they offer limited spatial–temporal consistency.

% (c) Motivated by these limitations, we propose a VLM-guided latent-space risk detector. Images are encoded into a latent world model, and semantic risk labels are obtained from a vision-language model. This combination leverages the predictive capability of the latent world model together with the semantic understanding of VLMs, achieving a favorable balance between computational efficiency, spatiotemporal reasoning and semantic understanding.

\begin{figure*}[ht]
    \centering
    \includegraphics[width=0.9\linewidth]{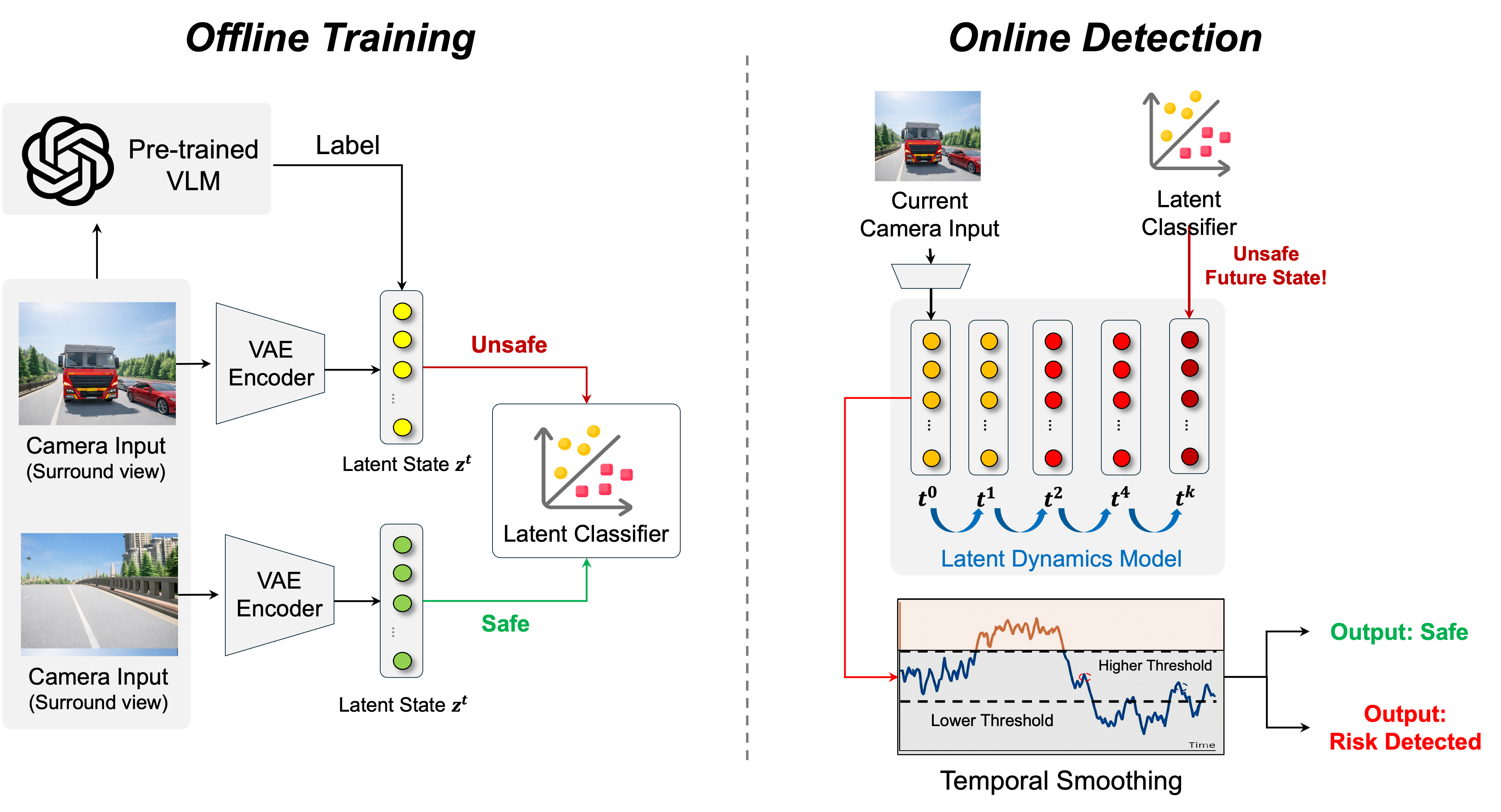}
    \caption{Overall pipeline of LSRE. 
A pretrained vision--language model (VLM) provides sparse semantic-risk
supervision for key frames.  
A recurrent state-space world model encodes multi-view observations into 
latent states with temporal dynamics, and generates short-horizon rollouts.  
A lightweight latent classifier, trained under VLM supervision, evaluates 
both instantaneous and predicted future latent states to produce a 
real-time semantic risk signal for the driving stack.}
    \label{fig:framework}
\end{figure*}

\section{Related Work}

\subsection{Safety Constraints and Shields in Autonomous Driving}
% Rules/heuristics, STL/temporal logic, reachability, safety critics, learning-based shields.
Early autonomous driving systems relied mainly on hand-crafted rules and deterministic state machines to encode traffic laws and basic driving behaviors~\cite{urmson2008autonomous}. These approaches work in structured environments but fail in ambiguous or  context-dependent scenarios. To obtain formal correctness, temporal-logic frameworks such as Linear Temporal Logic (LTL) and Signal Temporal Logic (STL) were introduced to specify and monitor safety constraints~\cite{kress2009temporal,donze2010robust}, and later extended to driving-related planning tasks~\cite{Sadigh-RSS-16}. 

Reachability analysis provided mathematically rigorous guarantees for collision avoidance. Hamilton--Jacobi (HJ) reachability computes forward reachable sets of unsafe states~\cite{1463302}, and later extensions introduced efficient approximations that support real-time reachability analysis in multi-agent interactive driving~\cite{9197129}. Control Barrier Functions (CBFs) were formalized into real-time enforceable safety constraints through quadratic program controllers for continuous systems~\cite{7782377}. 

Safe reinforcement learning introduced learning-based mechanisms to constrain policies during execution. Surveys highlight techniques such as constrained MDPs and safety critics that predict unsafe outcomes~\cite{garcia2015comprehensive}. Shielding approaches~\cite{jansen2020safe} monitor the agent’s actions and override unsafe actions before execution. 

Although effective for preventing collisions or boundary violations, these methods rely on explicit constraints and remain unsuitable for ambiguous or socially defined semantic contexts\cite{zhou2023identify}. Recent studies emphasize that many critical driving behaviors, such as yielding to emergency vehicles, prioritizing officers over traffic lights, or identifying hazards under occlusion, cannot be fully formalized using geometric rules or logical templates~\cite{shalev2017formal}. These limitations motivate our approach, which leverages high-level semantic supervision from VLMs to train a lightweight latent-space safety classifier.

\subsection{VLM-Based Risk Assessment and Driving Semantics}
% VLM for scene understanding, risks, traffic semantics; pros/cons for real time.

Vision–language models (VLMs) such as CLIP~\cite{radford2021learning}, BLIP-2~\cite{li2023blip}, and GPT-4V~\cite{yang2023dawn} have recently shown strong capabilities in visual reasoning and semantic understanding.
By aligning image and text representations, these models can interpret high-level scene semantics and contextual relationships beyond purely geometric features. Recent studies have applied VLMs to autonomous driving for scene understanding, captioning, and decision explanation~\cite{tian2024drivevlm,sima2023drivelm}, and further extended them to risk reasoning and intent recognition in interactive driving~\cite{kong2025vlr}.

However, direct use of VLMs in autonomous driving remains challenging due to high computational cost. Although model distillation and lightweight multimodal variants can partially mitigate these issues~\cite{yang2025edge,hegde2025distilling}, the resulting systems still require large-scale inference and lack strict runtime guarantees~\cite{zhou2024vision}. Moreover, VLM reasoning is often conducted on isolated images, making it difficult to maintain temporal coherence or safety-critical consistency~\cite{cui2024survey}. These limitations motivate us to explore how to retain the semantic understanding capabilities of VLMs while significantly improving runtime efficiency for real-time safety monitoring.

\subsection{World Models for Driving and Latent Monitoring}
% Dreamer-like RSSM, latent prediction; monitoring in latent space for safety.
World models aim to learn compact latent dynamics that capture environment transitions for imagination-based planning and decision making. Ha and Schmidhuber popularized a modern deep-learning formulation of  world models by learning recurrent latent dynamics that support  model-based reinforcement learning directly from pixels~\cite{ha2018world}. Later, the Dreamer series~\cite{hafner2023mastering} improved stability, scalability, and performance by introducing stochastic latent representations and actor-critic learning in latent space. These approaches demonstrated strong data efficiency and generalization in continuous-control tasks.

Recently, several driving frameworks have incorporated world-model concepts. DriveDreamer~\cite{wang2024drivedreamer} and DriveWorld~\cite{Min_2024_CVPR} use latent imagination to predict multi-modal driving behaviors and future risks. Such models enable representation learning that captures temporal context beyond single frames, forming a foundation for safety reasoning in latent space\cite{deng2021decision}.

However, existing methods mainly address geometric or stochastic risk, rather than high-level semantic safety. To address this, we propose a VLM-guided latent semantic risk estimator that transfers VLM-level understanding into a world-model latent space, enabling real-time semantic risk inference.

\section{Problem Formulation}
We consider an autonomous driving system with policy $\pi$ that generates control actions $a_t = \pi(o_t)$ from sensor observations $o_t$. Beyond geometric safety, the vehicle must comply with \textit{semantic safety constraints}—context-dependent human rules that dictate whether a state is socially acceptable or unsafe. Examples include yielding to emergency vehicles, following directions in temporary construction-zone, or stopping for a stopped school bus. These semantics are intuitive for humans but difficult to encode as explicit rules or logic.

Let $y_t \in \{0,1\}$ denote whether the driving state at time $t$ violates a semantic
constraint, and let $r_t \in [0,1]$ denote the probability of such a violation. The
objective is to learn a real-time semantic risk function
\begin{equation}
    r_t = g_\phi(o_t),
\end{equation}
which predicts whether the current observation is semantically unsafe.

\section{Method}

\subsection{System Overview}
% In this section, we describe the components of our proposed semantic risk detection framework. Our goal is to assess semantic safety in real time by combining high-level supervision from a vision-language model with the temporal reasoning capability of a latent world model. 

% Firstly, we introduce a VLM-guided supervision strategy that provides semantic annotations without manual labeling effort, which can provide the risk label for the following training.

% Secondly, we describe a lightweight latent-space risk classifier, including a world model (Recurrent State-Space Model) and a classifier. With a sufficient clips of driving trajectories, the world model can learn to autoregressively propagate latent states and capture the underlying dynamics of the environment. The surround-view observations are encoded into a compact latent space, within which the model is capable of generating forward rollouts that reflect plausible future states. Given the semantic annotations produced by the VLM, the classifier is subsequently trained to assign risk labels to these latent representations, enabling semantic assessment directly within the learned latent dynamics.

% Together, these components allow the system to integrate semantic understanding, temporal consistency, and computational efficiency within a unified architecture.

LSRE aims to detect semantic safety violations in real time by encoding 
language-defined rules into a latent space learned by a recurrent world model.  
The core idea is to use a VLM only as an \emph{offline semantic supervisor}—to extract sparse semantic labels—while performing all online inference through a lightweight latent classifier.

The framework consists of two main modules: (1) a VLM-guided semantic supervision mechanism that provides weak semantic labels for a small number of key frames; and (2) a semantic scoring module built on top of a recurrent state-space world model, containing both an instantaneous margin classifier and a short-horizon latent rollout value estimator. Together, these form a deployable semantic safety layer with temporal consistency and millisecond-level inference latency.

\subsection{VLM-Guided Semantic Supervision}

To obtain semantic supervision without relying on manual annotation, we employ a pretrained vision--language model (VLM) to generate pseudo-labels for a sparse set of key frames.  The driving sequence can be $\{x_t\}_{t=0}^{T}$, where each $x_t$ consists of four synchronized surround-view images and the ego-vehicle state.  

Every ten-frame segment is selected as a key frame and processed by the VLM to determine whether the scene contains a semantic safety risk.  Using a fixed prompt template, the VLM produces a soft semantic-risk label:
\begin{equation}
    \hat{y}_t = 
    \mathrm{VLM}\big(x_t^{(1)}, x_t^{(2)}, x_t^{(3)}, x_t^{(4)}, s_t; p\big),
    \qquad t \in \mathcal{K}
\end{equation}
where $x_t^{(i)}$ denotes the four-view images, $s_t$ the ego state, and $p$ the prompt.

Because semantic context typically varies slowly over short horizons, we assume that the semantic risk remains approximately stable within each ten-frame window:
\begin{equation}
    y_{t+k} \approx y_t, 
    \qquad k = 1,\ldots,9.
\end{equation}
To maintain temporal alignment across sparsely sampled key frames, we record the accumulated ego-motion over the skipped frames,
\begin{equation}
    \Delta s_{t^{-}\!\rightarrow t} = s_t - s_{t^{-}},
\end{equation}
where $t^{-}$ denotes the previous key frame.  These motion features serve as auxiliary information for the frames that are not directly processed by the VLM.  Along with the semantic decision obtained at the previous key frame $\hat{y}_{t^{-}}$, the accumulated ego-motion is supplied as additional input to the VLM when analyzing the subsequent key frame. This design compensates for missing intermediate observations and helps preserve temporal coherence across the reduced VLM inference frequency.

With these additional elements, the VLM query for each key frame is defined as

\begin{equation}
\begin{aligned}
    \hat{y}_t
    &= \mathrm{VLM}\!\Big(
        x_t^{(1)}, x_t^{(2)}, x_t^{(3)}, x_t^{(4)},\;
        s_t,\;
        \Delta s_{t^{-}\!\rightarrow t},\;
        \hat{y}_{t^{-}},\;
        p
    \Big), \\
    &\qquad\qquad\qquad\quad t \in \mathcal{K}
\end{aligned}
\end{equation}
where $x_t^{(i)}$ denotes the four synchronized surround-view images, $s_t$ is the current ego-vehicle state, 
$\Delta s_{t^{-}\!\rightarrow t}$ is the accumulated motion across skipped frames, $\hat{y}_{t^{-}}$ is the semantic judgment from the previous key frame, and $p$ is the fixed prompt template. This extended formulation ensures that the supervisory signal reflects short-term scene evolution despite the sparse sampling of VLM queries.

\subsection{Semantic Risk Scoring Module}

LSRE performs semantic safety assessment inside the latent space of a 
recurrent state-space model (RSSM).  
The RSSM provides a compact latent representation $z_t$ with short-horizon
temporal consistency, while all semantic reasoning is handled by a lightweight
classifier trained under VLM supervision.

\subsubsection{Latent dynamics}
Given observation $o_t$, the encoder (inference model) infers a posterior latent
\begin{equation}
  z_t \sim q_\psi(z_t \mid \hat z_t, o_t),  
\end{equation}

where $\hat z_t$ denotes the prior (predicted) latent before incorporating $o_t$.
The transition model propagates the latent state forward as
\begin{equation}
\hat z_{t+1} \sim p_\phi(\hat z_{t+1}\mid z_t, a_t),
\end{equation}
providing predictive structure without requiring additional semantic supervision.
% Given observation $o_t$, the encoder produces a posterior latent state 
% $ z_t\sim q_\psi(z_t\mid \hat z_t, o_t)$ together with a latent state 
% $\hat z_t$.  
% The transition model propagates the latent state forward as
% \begin{equation}
% \hat z_{t+1}\sim p_\phi(\hat z_{t+1}\mid z_t,a_t),
% \end{equation}
% providing predictive structure but introducing no new semantic information.

\subsubsection{Margin-Based Semantic Risk Classifier}
To evaluate whether a latent state violates a semantic safety constraint, we train a classifier $g_\mu(z_t)$ on top of the RSSM latent representation.  The classifier outputs a real-valued margin score, where large positive values indicate safe states and large negative values indicate semantic violations.

Given training sets $D$, with a signed label $y_i \in \{0,1\}$ indicating safe or unsafe samples, the classifier is trained to satisfy a signed margin constraint, where violations of $y_i g_\mu(z_i) \ge \delta$ are penalized as follows:

\begin{equation}
\mathcal{L}_\mu
= \frac{1}{N}
\sum_{z_i \in D}
\mathrm{ReLU}\!\Big(\,
\delta - y_i\, g_\mu(z_i)
\Big),
\label{eq:margin_loss_alt}
\end{equation}

This loss encourages separation between safe and unsafe latent states in a geometrically interpretable way, providing an instantaneous semantic risk indicator.

\subsubsection{Future Semantic Value Estimation}

While $g_\mu(z_t)$ provides an instantaneous estimate of the semantic risk at the current frame, it does not reflect violations that may emerge in the near future.  To capture such prospective risks, we estimate a latent value function $V_{\mathrm{latent}}(z_t)$ by rolling out the world model for a fixed horizon of $K=50$ steps and accumulating the predicted margin values along the simulated trajectory.

Starting from the current latent state $z_t$, the RSSM transition model generates future latent states,
\begin{equation}
    z_{t+k} \sim p_\phi\!\left( \,\cdot \mid z_{t+k-1}, a_{t+k-1} \right),
    \qquad k = 1,\ldots, K,
\end{equation}
where $a_{t+k-1}$ denotes the ego action applied during rollout.

The latent value is computed as the discounted sum of future margin scores:
\begin{equation}
\begin{aligned}
    V_{\mathrm{latent}}(z_t)
    &= \sum_{k=0}^{K}
       \gamma^{k}\,g_\mu(z_{t+k}), \\
    &\qquad\qquad  0 < \gamma \le 1,
\end{aligned}
\label{eq:discounted_rollout}
\end{equation}
where $\gamma$ is the discount factor controlling the contribution of future risk.

This finite-horizon approximation yields a conservative estimate of semantic risk reachable within the next 50 frames, enabling the system to detect safety violations that may not yet be visible at the current time.

\subsubsection{Hysteresis-Based Temporal Smoothing}

The margin score $g_\mu(z_t)$ may fluctuate due to uncertainties in the latent representation, especially near the decision boundary.  To avoid spurious state transitions caused by these small oscillations, we adopt a hysteresis thresholding strategy~\cite{renevey2001entropy}.  This mechanism introduces two thresholds---an upper threshold $\theta_{\mathrm{high}}$ and a lower threshold $\theta_{\mathrm{low}}$ with $\theta_{\mathrm{low}} < \theta_{\mathrm{high}}$ , such that the semantic state is updated only when the  margin crosses either threshold.

Let $y_t \in \{0,1\}$ denote the binary semantic risk indicator.  The hysteresis rule is defined as:
\begin{equation}
y_t =
\begin{cases}
0, & g_\mu(z_t) \ge \theta_{\mathrm{high}}, \\[4pt]
1, & g_\mu(z_t) \le \theta_{\mathrm{low}},  \\[4pt]
y_{t-1}, & \text{otherwise},
\end{cases}
\label{eq:hysteresis}
\end{equation}
where the third case preserves the previous state when the margin lies within the hysteresis band $[\theta_{\mathrm{low}},\,\theta_{\mathrm{high}}]$.

This design prevents rapid switching between safe and unsafe predictions and yields a more stable semantic risk signal for downstream decision making.

\section{Experiment}

\subsection{Experiment Setup}

\subsubsection{Test Scenarios}
\begin{figure}[t]
    \centering
    \includegraphics[width=1\linewidth]{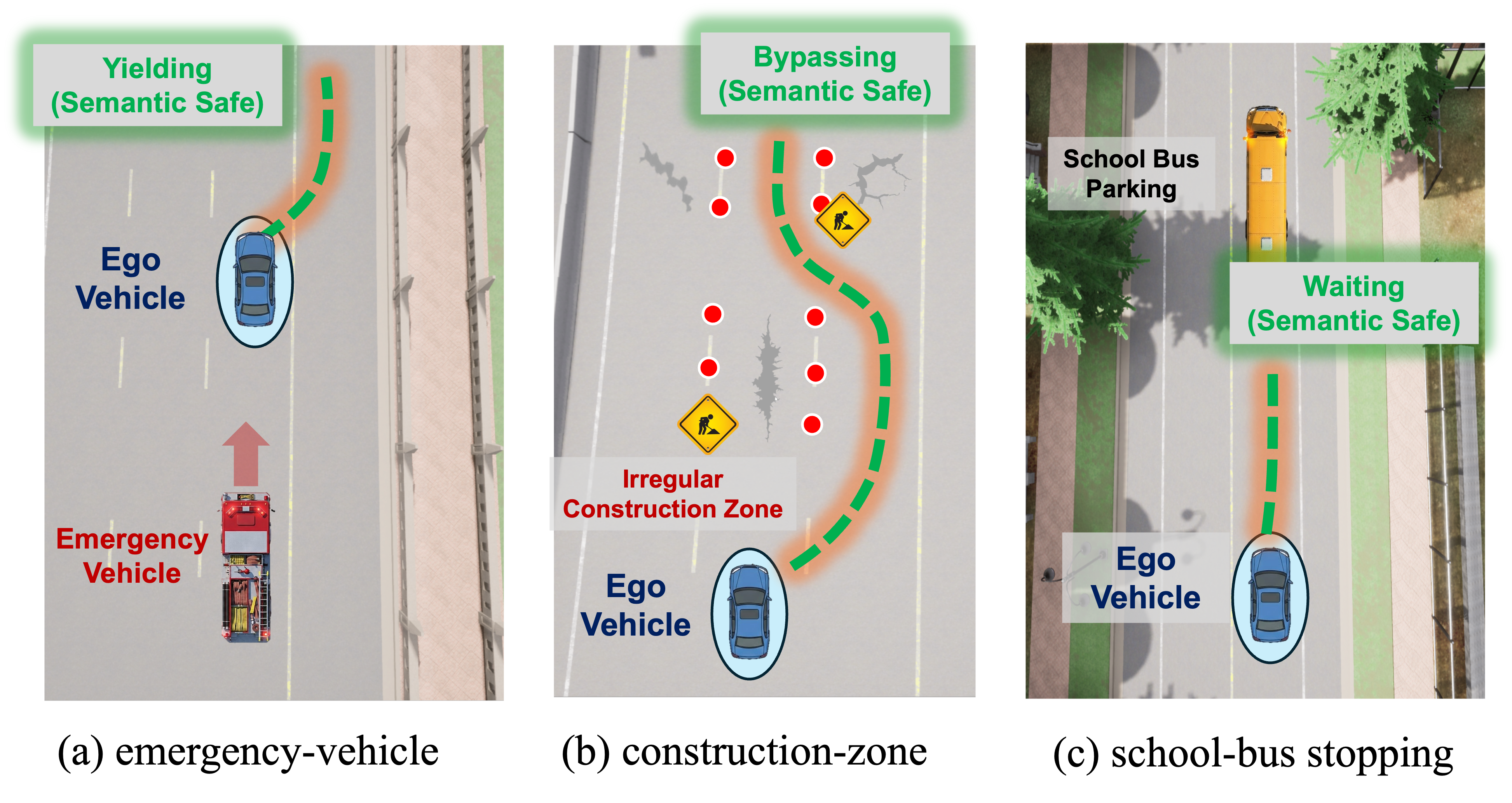}
    \caption{Three semantic–failure categories used in evaluation, each instantiated with two scenario variants (in–distribution and few–shot).
These scenarios capture human-understandable but rule-hard safety semantics that are essential for evaluating real-time semantic risk detection.}
    \label{fig:test_cases}
\end{figure}

We evaluate the proposed semantic safety framework across six representative scenarios in CARLA, as shown in Fig.\ref{fig:test_cases}\cite{qian2024spider}. Our benchmark contains three core semantic–failure categories, each instantiated with two scenario variants (in–distribution and few–shot):  (1) rear-approaching emergency-vehicle yielding, (2) construction-zone lane inference, and (3) school-bus stopping.  Each category contains two sub-settings designed to assess both in-distribution performance and semantic generalization.
These variants share the same semantic rule but differ significantly in appearance and layout, forming a total of six semantic-risk scenarios.

\paragraph{In-distribution scenario}
For each scenario category, we construct 100 diverse simulation clips (10\,s at 10\,Hz) covering variations in road geometry, traffic density, and object appearance.  These clips are used for training, validation, and testing within the same category to measure frame-wise semantic risk detection under sufficient data coverage.

\paragraph{Semantically similar few-shot scenario}
To evaluate semantic rather than spatial memorization, we additionally construct a semantically similar but visually and geographically distinct variant of each scenario, using only 10 clips for training.  

The corresponding test set contains 100 unseen clips that differ in layout, lighting, and agent configurations while preserving the same semantic rule  (e.g., yielding to an approaching emergency vehicle, considering construction bottlenecks, or stopping for a school bus).  Performance on this setting reflects the model’s ability to transfer semantic safety rules under limited supervision and unseen environments.

\subsubsection{Baselines}

\paragraph{VLM‐Only}
As a reference for semantic risk detection, we include a direct VLM baseline using the pretrained GPT-5-mini model. For each input frame, the VLM produces a binary semantic-risk judgment indicating whether the current scene violates any semantic safety constraints. This baseline reflects the performance of a high-capacity model performing per-frame semantic interpretation, and serves as a comparison point for our lightweight classifier built on top of the learned world model. In our setup, the VLM is accessed via a cloud API, and the reported runtime includes both model inference and network communication latency.

\paragraph{Always‐Safe}

We additionally include a trivial ``Always-Safe'' baseline, which always predicts every frame as semantically safe.  Although simplistic, this baseline provides a useful lower bound by illustrating the performance of a model that never flags any violation.  It allows us to quantify how much improvement is gained by incorporating semantic supervision and latent-state reasoning, particularly on metrics that depend on detecting positive risk cases such as recall and accuracy score.

\subsubsection{Evaluation Metrics}

We evaluate the semantic‐safety detection performance using three primary metrics: Accuracy, Recall, and False‐Alarm Rate (FAR).   Accuracy measures the overall correctness of frame‐level predictions: \begin{equation} \text{Acc} = \frac{TP + TN}{TP + TN + FP + FN}.
\end{equation}  Recall reflects the proportion of true unsafe frames that are correctly identified: \begin{equation} \text{Rec} = \frac{TP}{TP + FN}.
\end{equation}  
Latency is reported as the median and 95th percentile of the end‐to‐end decision delay,  indicating real‐time feasibility under vehicle‐level execution constraints.  We present both frame‐level (micro) results over 60\,000 frames and  scenario‐level (macro) averages across the six semantic‐failure categories.  For each semantic-failure event, we measure how early the method raises an unsafe signal relative to the annotated event onset. This metric reflects the model’s ability to anticipate upcoming hazards rather than reacting only after the risky situation fully materializes. 

We also quantify the proportion of erroneous unsafe predictions during normal routine driving with no semantic failures. This monitors stability and practicality during daily deployment, ensuring the semantic-safety module does not produce disruptive or distracting alerts. The False‐Alarm Rate (FAR) quantifies the ratio of normal frames that are mistakenly flagged as unsafe:
\begin{equation} \text{FAR} = \frac{FP}{FP + TN}.
\end{equation}  
Together, these metrics characterize not only instantaneous correctness but also temporal anticipation and deployment robustness, providing a comprehensive evaluation of semantic-safety performance.

\begin{table}[t]
\centering
\renewcommand{\arraystretch}{1.22}
\caption{Frame-wise semantic risk detection across three semantic-failure 
categories under in-distribution and few-shot settings.}
\label{tab:scenario_two_setting_clean}
\begin{tabular}{lcccc}
\toprule
\multirow{2}{*}{\textbf{Category / Method}} &
\multicolumn{2}{c}{\textbf{In-dist Scenario}} &
\multicolumn{2}{c}{\textbf{Few-shot Scenario}} \\
\cmidrule(lr){2-3}
\cmidrule(lr){4-5}
& \textbf{Acc (\%)} & \textbf{Rec (\%)} 
& \textbf{Acc (\%)} & \textbf{Rec (\%)} \\
\midrule

\multicolumn{5}{l}{\textbf{Category~1: Emergency Vehicle}} \\
\hspace{0.7em} Always-Safe  
& 54.36 & 0.00 
& 66.64 & 0.00 \\
\hspace{0.7em} VLM-Only    
& 91.80 & 84.07
& 93.38 & 98.53 \\
\hspace{0.7em} LSRE (Ours)         
& 85.66 & 98.64
& 91.29 & 92.06 \\
\midrule

\multicolumn{5}{l}{\textbf{Category~2: Construction Zone}} \\
\hspace{0.7em} Always-Safe  
& 33.49 & 0.00 
& 42.03 & 0.00 \\
\hspace{0.7em} VLM-Only    
& 93.49 & 90.36
& 79.40 & 74.36 \\
\hspace{0.7em} LSRE (Ours)         
& 89.54 & 91.61
& 72.57 & 76.90 \\
\midrule

\multicolumn{5}{l}{\textbf{Category~3: School Bus}} \\
\hspace{0.7em} Always-Safe  
& 27.60 & 0.00
& 59.79 & 0.00 \\
\hspace{0.7em} VLM-Only    
& 97.14 & 96.71
& 81.69 & 96.24 \\
\hspace{0.7em} LSRE (Ours)         
& 93.32 & 93.08
& 79.88 & 93.14 \\
\midrule

\textbf{Overall Avg.} \\
\hspace{0.7em} Always-Safe  
& 38.48 & 0.00 
& 56.15 & 0.00 \\
\hspace{0.7em} VLM-Only    
& 94.14 & 90.38
& 84.82 & 89.71 \\
\hspace{0.7em} LSRE (Ours)         
& 89.51 & 94.44
& 81.25 & 87.37 \\
\bottomrule
\end{tabular}
\end{table}

\begin{table*}[ht]
\centering
\renewcommand{\arraystretch}{1.25}
\caption{Event-level early warning analysis across three semantic-failure categories.
Event Recall measures whether an event is detected at least once; 
Average Lead Time indicates anticipation relative to the annotated onset.}
\label{tab:event_early_warning_all}
\begin{tabular}{lcccccccc}
\toprule
\multirow{2}{*}{\textbf{Method}} &
\multicolumn{2}{c}{\textbf{Emergency Vehicle}} &
\multicolumn{2}{c}{\textbf{Construction Zone}} &
\multicolumn{2}{c}{\textbf{School Bus Stop}} &
\multicolumn{2}{c}{\textbf{Overall}} \\
\cmidrule(lr){2-3}
\cmidrule(lr){4-5}
\cmidrule(lr){6-7}
\cmidrule(lr){8-9}
& Recall (\%) & Lead (ms) &
  Recall (\%) & Lead (ms) &
  Recall (\%) & Lead (ms) &
  Recall (\%) & Lead (ms) \\
\midrule
Heuristic     & 100.0\%  & 0 & 100.0\%  & 0 &100.0\%  & 0 & 100.0\%  & 0 \\
VLM-Only      & 97.0\% & 51.5 & 100.0\%  & 248.0 & 100.0\% &454.5 & 99.0\% & 51.7 \\
\textbf{LSRE (Ours)} & \textbf{99.5\%} & \textbf{1480.0}
              & \textbf{100.0\%} & \textbf{3273.5}
              & \textbf{100.0\%} & \textbf{2792.5}
              & \textbf{99.8\%} & \textbf{2515.3} \\
\bottomrule
\end{tabular}
\end{table*}

\subsection{Frame-wise Risk Detection in Long-tail Scenarios}

We first evaluate frame-level semantic risk detection on our dataset, which contains six categories of long-tail semantic failures including emergency-vehicle interactions, temporary construction layouts, and school-bus stops. Each category includes 100 clips (10\,s at 10\,Hz), yielding 60\,000 frame-wise safe/unsafe labels. This experiment isolates the per-frame classification problem to examine how well each method identifies semantic violations without relying on temporal smoothing or event structure.

The primary goal of this evaluation is to measure how faithfully a model can reproduce the semantic judgments of a VLM. Since LSRE distills VLM outputs into a lightweight classifier, strong frame-wise performance indicates successful semantic rule encoding. For each method, we compute micro-level Accuracy and Recall across all frames: Accuracy reflects overall correctness, while Recall quantifies the ability to avoid missing risky frames—critical for semantic safety monitoring. Together, these metrics assess the quality of real-time semantic safety signals that would be injected into the driving stack.

Table~\ref{tab:scenario_two_setting_clean} reports results for two evaluation settings: in-distribution and few-shot.

\paragraph*{In-distribution scenario}
Across the three categories, our method achieves an overall accuracy of 89.51\%, close to the VLM-only baseline (94.14\%). Recall reaches 94.44\%, exceeding the VLM-only model and indicating strong preservation of VLM semantic sensitivity under matched distribution conditions.%However, due to its conservative design, our method yields a modestly higher false-alarm rate 15.32\% overall compared with 1.25\% for the VLM-only baseline, reflecting a tendency to flag ambiguous frames as unsafe.

\paragraph*{Few-shot scenario}
When supervision is extremely limited, performance differences become more pronounced. Even so, our method maintains 81.25\% accuracy compared with the VLM-only baseline (84.82\%), demonstrating that the semantic rules distilled into LSRE remain robust under sparse supervision.
  %The conservative nature of our classifier again leads to a higher FAR (27.64\% vs.\ 12.57\% for the VLM-only model), but this trade-off enables the system to avoid missing true risky frames, which is especially critical in low-data regimes.

\paragraph*{Summary}
Across both settings, LSRE achieves frame-level accuracy and recall comparable to the large-model baseline, confirming that the latent classifier effectively captures VLM-level semantic capability while operating with significantly lower inference overhead.

\subsection{Event-level Early Warning Analysis}

While frame-wise classification evaluates semantic understanding on isolated observations, real autonomous driving requires anticipating hazardous situations before they fully unfold. Event-based evaluation therefore provides a more realistic assessment of a system’s deployability: a semantic safety module should not only identify unsafe frames, but also predict the onset of safety-critical events in advance.

For each semantic-failure category, we manually annotate the onset time of the hazardous event, such as the moment an emergency vehicle first becomes visible within a 30\,m range, the emergence of a construction bottleneck, or the instant a school bus activates its stop sign. An early warning is counted as correct if the model raises an unsafe flag at any time between the annotated onset and the end of the event. For all correctly detected events, we measure the lead time, defined as the time difference between the first predicted unsafe frame and the annotated onset.
 
We report the following event-level metrics:

\paragraph{Event Recall} percentage of risky events that are detected at least once;

\paragraph{Average Lead Time} average anticipation time across detected events.

% Table~\ref{tab:event_early_warning_all} presents the event-level early-warning results across the three scenarios. Both methods achieve high event recall, consistently detecting every risky event. However, our approach raises the unsafe flag substantially earlier than the VLM-only baseline across all categories, thanks to the latent value function for future semantic risk. Compared with the VLM-only baseline, our method triggers warnings much earlier—for instance, the average lead time is \textbf{3273.5\,ms} versus \textbf{248.0\,ms} in Construction Zone, and \textbf{2792.5\,ms} versus \textbf{454.5\,ms} in School Bus, showing a clear advantage in anticipating the onset of semantic hazards. Overall, our method increases the average lead time by a large margin while maintaining the same high recall, demonstrating that the latent classifier can recognize safety-critical semantics much earlier and provide more timely warnings than the large-model baseline.

Table~\ref{tab:event_early_warning_all} summarizes the results. Both methods achieve nearly identical event recall, consistently detecting every hazardous event. However, our approach provides substantially earlier warnings than the VLM-only baseline across all categories. This improvement stems from the temporal predictive structure of the world model, which enables the latent classifier to recognize future semantic violations before they are visually obvious. For example, the average lead time in the Construction Zone scenario is 3273.5\,ms for our method compared with 248.0\,ms for the VLM-only baseline, and 2792.5\,ms versus 454.5\,ms in the School Bus scenario. These results demonstrate that LSRE offers significantly earlier anticipation while maintaining high recall, making it more suitable for real-time semantic safety monitoring in autonomous driving.

\subsection{False Alarms in Normal Driving Scenario}

To evaluate the stability of the proposed semantic-safety module during everyday operation, we test all methods on 30 minutes of normal driving in CARLA, covering routine behaviors such as car-following, lane-keeping, signal compliance, and non-critical interactions with pedestrians and vehicles. No semantic-failure events occur in these logs, so any unsafe prediction is counted as a false alarm.

We report the frame-wise False-Alarm Rate (FAR), averaged across all normal-driving segments. Maintaining a low FAR is crucial for real-world deployment because excessive alarms may destabilize downstream planning modules, cause unnecessary slowdowns, and lead to driver disengagements.

Table~\ref{tab:false_alarm_normal} shows that the latent classifier alone produces occasional spurious unsafe predictions during routine driving, yielding an 8.29\% FAR due to fluctuations and uncertainties in latent features. After applying the proposed hysteresis-based temporal filtering mechanism, the FAR is reduced to 0.98\%, eliminating nearly all false triggers. This reduction highlights the importance of temporal filtering: with the filter applied, the semantic-safety signal becomes sufficiently stable for practical integration into an autonomous driving stack.

\begin{table}[t]
\centering
\renewcommand{\arraystretch}{1.15}
\caption{False-alarm rate in normal driving scenarios (no semantic failures). Lower is better.}
\label{tab:false_alarm_normal}
\begin{tabular}{lc}
\toprule
\textbf{Method} & \textbf{FAR (\%)} \\
\midrule
LSRE w/o filter & 8.29 \\
LSRE           & 0.98 \\
\bottomrule
\end{tabular}
\end{table}

\subsection{Real-time Performance}
Beyond detection accuracy, a semantic-safety module must satisfy the timing constraints of an autonomous driving stack, where perception and planning typically operate at 10--20~Hz. We therefore evaluate the end-to-end inference latency of all methods.

Latency is measured as the elapsed time between receiving a new frame and outputting a semantic-risk decision. We repeat each method for 100 queries and report both the median and the 95th-percentile (p95) to characterize typical and tail latency.

Experiments are conducted on a workstation running Ubuntu~22.04 with an Intel Xeon Gold 5218R CPU, 256~GB RAM, and an NVIDIA Tesla GV100 GPU, using batch size 1 to reflect streaming, per-frame inference. For \textit{VLM-Only}, we query a server-scale VLM via a cloud API---a practical choice given that such models are often infeasible to deploy on-board---and thus the reported latency includes both model inference and network communication overhead. In contrast, \textit{LSRE} runs fully on-device.

As shown in Table~IV, \textit{VLM-Only} is far from real-time, requiring multi-second end-to-end latency (2917~ms median, 3706~ms p95). In contrast, \textit{LSRE} achieves 9.44~ms median and 11.91~ms p95, corresponding to over a 300$\times$ speedup and comfortably meeting the real-time budget for 10--20~Hz operation. These results suggest that LSRE retains VLM-level semantic guidance through offline supervision while enabling lightweight, real-time on-vehicle deployment.
% Beyond detection accuracy, a semantic-safety module must satisfy the timing constraints of an autonomous driving stack, where perception and planning typically operate at 10--20\,Hz. We therefore evaluate the end-to-end inference latency of all methods under a unified hardware setup. Latency is measured as the time between receiving a new frame and producing a semantic-risk decision over 100 repeated queries, and we report both median and 95th-percentile (p95) latency to capture typical and worst-case behavior.

% Table~\ref{tab:latency_table} shows that the VLM-only baseline is far from real-time operation, requiring multi-second inference (2917\,ms median, 3706\,ms p95) due to large-model processing. In contrast, our latent classifier runs in 9.44\,ms median latency and 11.91\,ms p95, achieving over a 300$\times$ improvement. This places the entire semantic-safety module comfortably within the control-cycle budget of modern autonomous vehicles.

% These results indicate that LSRE preserves the high-level semantic guidance of the VLM while meeting strict real-time requirements. Unlike the VLM-only approach, whose multi-second delay makes online deployment infeasible, the lightweight latent classifier achieves sub-20\,ms end-to-end latency at 10\,Hz, making it practical for integration into real on-vehicle systems.

\begin{table}[t]
\centering
\renewcommand{\arraystretch}{1.18}
\caption{End-to-end inference latency under a unified hardware setup. 
Median and 95th-percentile (p95) values are reported.}
\label{tab:latency_table}
\begin{tabular}{lcc}
\toprule
\textbf{Method} & \textbf{Median (ms)} & \textbf{p95 (ms)} \\
\midrule % <— 表头下横线
VLM-Only      & 2917 & 3706 \\
LSRE (Ours)   & 9.44 & 11.91 \\
\bottomrule
\end{tabular}
\end{table}

\subsection{Case Study}
% \begin{figure}
%     \centering
%     \includegraphics[width=1\linewidth]{images/casestudy-1230.jpg}
%    \caption{   Real-Time Outputs of LSRE in a Right-Turn Encounter}
%     \label{fig:idea}
% \end{figure}

\begin{figure}[t]
    \centering

    \begin{subfigure}[t]{\linewidth}
        \centering
        \includegraphics[width=\linewidth]{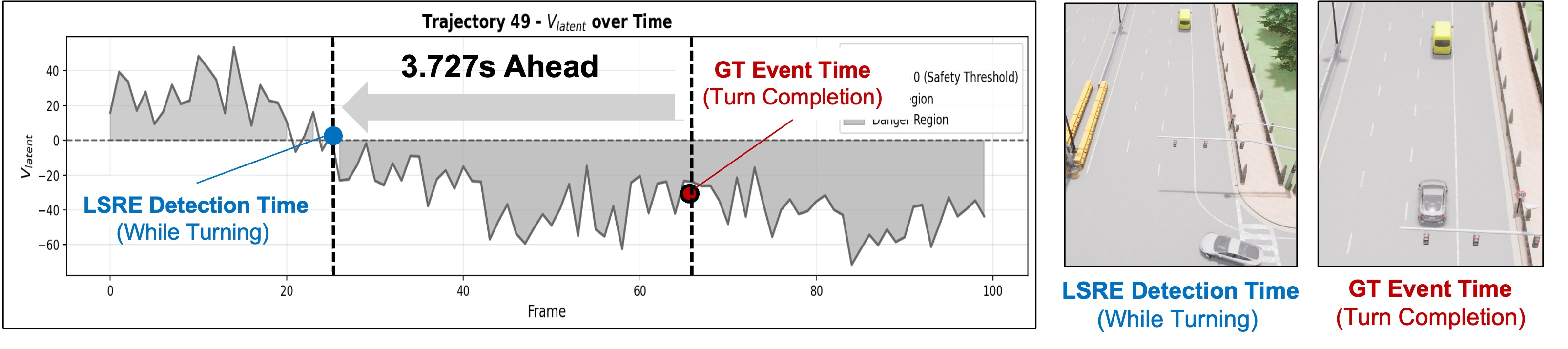}
        \caption{Case 1: School-Bus Stopping}
        \label{fig:cs_bus}
    \end{subfigure}

    \vspace{0.6em}

    \begin{subfigure}[t]{\linewidth}
        \centering
        \includegraphics[width=\linewidth]{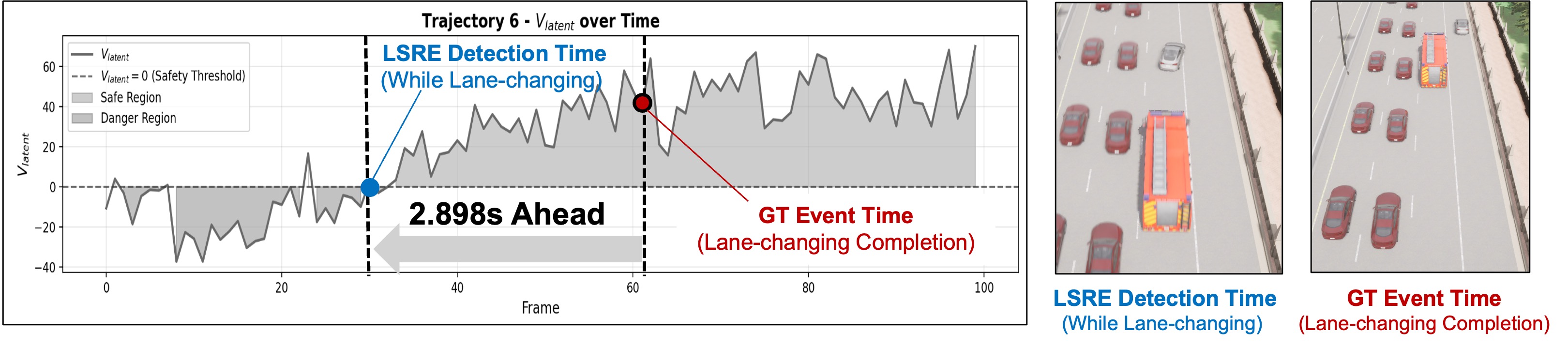}
        \caption{Case 2: Emergency-Vehicle Yielding}
        \label{fig:cs_emg}
    \end{subfigure}

    \vspace{0.6em}

    \begin{subfigure}[t]{\linewidth}
        \centering
        \includegraphics[width=\linewidth]{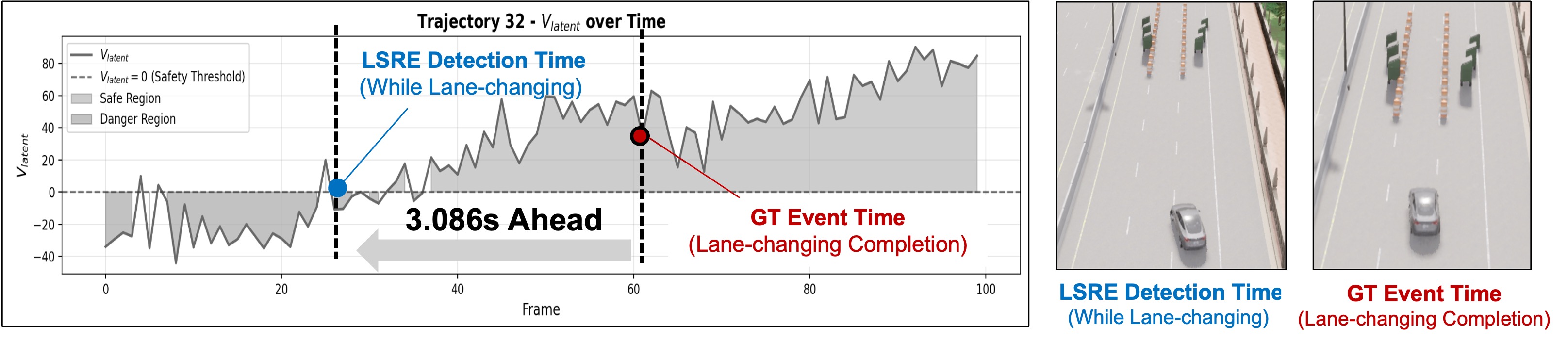}
        \caption{Case 3: Construction-Zone}
        \label{fig:cs_const}
    \end{subfigure}

    \caption{The semantic risk value output by the proposed LSRE model across three example cases. Each subplot highlights the LSRE detection time and the ground-truth (GT) risk happened time.}
    \label{fig:case_study}
\end{figure}

Fig. 3 visualizes the real-time signals produced by LSRE across three representative driving scenarios. The system outputs the future semantic value estimation $V_{\mathrm{latent}}$ over time. Dashed horizontal lines indicate the zero threshold ($V_{\mathrm{latent}}=0$, safe vs. unsafe), while the shaded backgrounds mark the inferred safe and danger regions. We highlight two key timestamps: the \emph{LSRE detection time}, defined as the first time $V_{\mathrm{latent}}$ crosses the threshold, and the \emph{ground-truth (GT) boundary time}, i.e., the annotated transition between semantic safe and unsafe states.

In an example case from School-Bus Stopping scenario, the ego vehicle initiates a right turn at an intersection and later encounters a stopped yellow school bus in its lane, resulting in a gradually developing hazard. Notably, $V_{\mathrm{latent}}$ starts to decrease at the \emph{beginning of the turning maneuver} and crosses the $V_{\mathrm{latent}}=0$ threshold \emph{well before} the GT boundary time, which occurs at turn completion. This yields a lead time of $3.727\,\mathrm{s}$. Similarly, in example cases from Emergency-Vehicle Yielding and Construction-Zone scenarios, LSRE identifies the annotated safety-state transition ahead of the GT boundary, with lead times of $2.898\,\mathrm{s}$ and $3.086\,\mathrm{s}$, respectively.

Overall, these examples show that $V_{\mathrm{latent}}$ provides a real-time estimate of the current semantic safety state, and can further anticipate upcoming safety transitions by several seconds relative to the GT boundary annotations, enabling timely semantic safety monitoring under online deployment constraints.

\section{Conclusion}
\label{section:conclusions}

In this paper, we introduced LSRE, a VLM-guided latent semantic rule encoding framework for real-time semantic risk detection in autonomous driving. By using a vision–language model solely as an offline semantic supervisor and deploying a lightweight classifier in the latent space of a recurrent world model, the proposed approach achieves an effective balance between semantic awareness and runtime efficiency. Experiments across six semantic-failure scenarios demonstrate that LSRE achieves detection accuracy comparable to a large VLM baseline, while offering substantially earlier event-level warnings and maintaining a low false-alarm rate in normal driving, all within sub-100\,ms end-to-end latency suitable for on-vehicle execution.

In future work, we plan to advance from semantic detection toward failure-aware response by coupling the learned risk signals with concrete fallback behaviors, such as conservative re-planning, safe deceleration, or handover strategies under classifier uncertainty. Another direction is to integrate the semantic-safety layer more tightly with downstream planning and control to provide closed-loop guarantees. Finally, we aim to evaluate LSRE on large-scale real-world driving data. These steps seek to further bridge language-informed semantic reasoning with the practical safety requirements of autonomous driving systems.

\bibliographystyle{ieeetr}
\bibliography{ref}

\end{document}